\documentclass{article}
\usepackage{spconf,amsmath,graphicx}
\usepackage{algorithmic}
\usepackage{algorithm}
\usepackage{multirow}
\usepackage{amsmath,amssymb}

\title{Learning Audio-Visual Correlations from Variational Cross-Modal Generation}
\name{Ye Zhu$^{1}$ \qquad Yu Wu$^{2}$ \qquad Hugo Latapie$^{3}$ \qquad Yi Yang$^{2}$ \qquad Yan Yan$^{1}$ }
  
\address{$^{1}$ Illinois Institute of Technology, USA \\
      $^{2}$ ReLER, University of Technology Sydney, Australia \\
      $^{3}$ Cisco, USA}

\begin{document}
%
\maketitle
\begin{abstract}
People can easily imagine the potential sound while seeing an event. This natural synchronization between audio and visual signals reveals their intrinsic correlations.
To this end, we propose to learn the audio-visual correlations from the perspective of cross-modal generation in a self-supervised manner, the learned correlations can be then readily applied in multiple downstream tasks such as the audio-visual cross-modal localization and retrieval. We introduce a novel Variational AutoEncoder (\emph{VAE}) framework that consists of Multiple encoders and a Shared decoder (\emph{MS-VAE}) with an additional Wasserstein distance constraint to tackle the problem. 
Extensive experiments demonstrate that the optimized latent representation of the proposed \emph{MS-VAE} can effectively learn the audio-visual correlations and can be readily applied in multiple audio-visual downstream tasks to achieve competitive performance even without any given label information during training.
\end{abstract}
\section{Introduction}
\label{sec:intro}

As humans, we can naturally imagine the possible visual frames while hearing the corresponding sound or imagine the potential sound while seeing an event happening.
Therefore, the correlations between audio and visual information accompanying an event can be modeled in the perspective of generation, \emph{i.e.}, the corresponding audio signals and visual frames can generate each other. Moreover, the natural correspondence between audio and visual information from the videos makes it possible to accomplish this objective in a self-supervised manner without additional annotations.

Audio and visual perceptions are both essential sources of information for humans to explore the world. Audio-visual cross-modal learning has thus become a research focus in recent years~\cite{korbar2018cooperative,senocak2018learning,arandjelovic2018objects,tian2019audio,owens2018audio,rouditchenko2019self,hu2019deep,gan2020music,gan2020foley}. The correlations between audio and visual signals are the key in this field. Recent studies in the audio-visual cross-modal field largely focus on representation learning that incorporates the information from both modalities in a discriminative way, and then applying the learned feature embedding in relevant audio-visual tasks such as sound source localization~\cite{owens2018audio,gao2018learning,zhao2018sound}, sound source separation~\cite{senocak2018learning,arandjelovic2018objects,zhao2018sound}, cross-modal retrievals~\cite{aytar2016soundnet,arandjelovic2018objects,owens2018audio,tian2018ave} and cross-modal localization~\cite{tian2018ave,wu2019DAM}. 
In contrast, another branch of research work exploits to learn the correlations in a generative manner~\cite{chen2017deep,rouditchenko2019self,rana2019towards}. Hu \textit{et al.}~\cite{hu2019deep} introduce Deep Multimodal Clustering for capturing the audio-visual correspondence.
Korbar \textit{et al.}~\cite{korbar2018cooperative} propose a cooperative learning schema to obtain multi-sensory representation from self-supervised synchronization.
Arandjelovic and Zisserman~\cite{arandjelovic2018objects} propose to learn a mutual feature embedding through the audio-visual correspondence task. As for more concrete audio-visual downstream tasks, Gao \textit{et al.}~\cite{gao2018learning} look into the problem of separating different sound sources based on a deep multi-instance multi-label network. 
Among those above studies with concrete downstream tasks, most of them learn the audio-visual correlations in a discriminative way to obtain better performance, which usually requires label information. Our work tackles the problem from a different perspective based on generative models via self-supervised learning.

\begin{figure*}[t]
    \centering
    \includegraphics[width=0.9\textwidth]{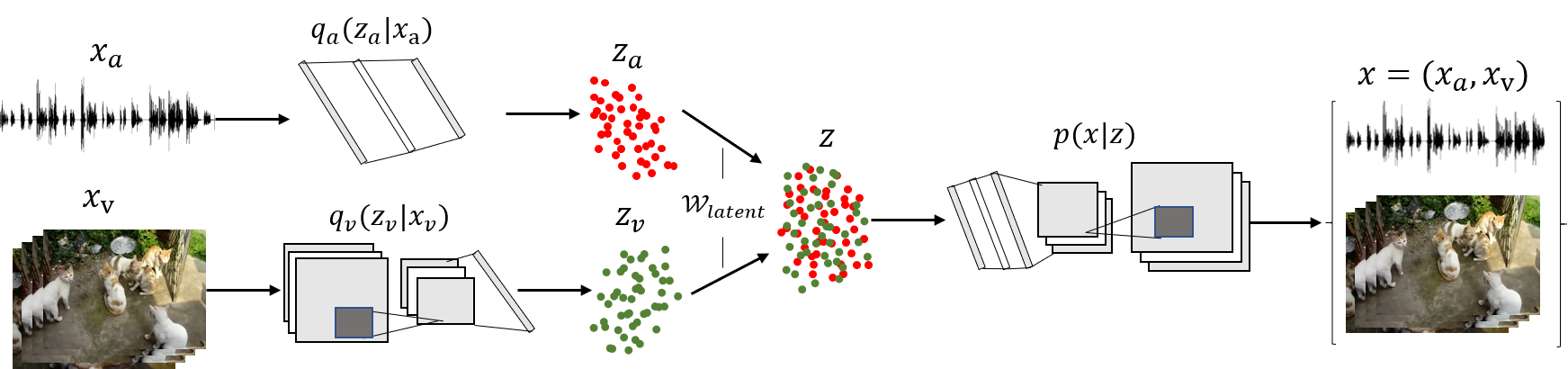}
    \caption{Schematic overview of our proposed \emph{MS-VAE} model. 
    }
    \label{fig:msvae}
\end{figure*}

 Overall, we have two motivations to fulfill in this work: leveraging the label-free advantage to learn the intrinsic audio-visual correlations from cross-modal generations via the proposed \emph{MS-VAE} framework, and achieving competitive performance for multiple audio-visual downstream tasks at the same time. 
 Specifically, the proposed \emph{MS-VAE} is a Variational AutoEncoder framework with Multiple encoders and a Shared decoder. 
VAE~\cite{kingma2013auto} is a popular class of generative models that synthesizes data with latent variables sampled from a variational distribution, generally modeled by Gaussian distributions. Based on the properties of VAE, we propose to use the latent variables to represent modality-specific data. Then these latent variables need to be aligned in order to complementarily present the happening event in two perspectives. Finally, the aligned latent variable should be able to generate both audio and visual information to construct the corresponding data pair. The optimized latent variable hence automatically learns the intrinsic audio-visual correlations during the process of cross-modal generation, and is ready to be directly applied in audio-visual tasks. 

One practical challenge to apply the VAE framework in multi-modal learning is that VAE often suffers from the degeneration in balancing simple and complex distributions~\cite{zheng2018degeneration} due to the large dimensionality difference between audio and video data. We adopt a shared-decoder architecture to help avoiding the degeneration and to enforce the mutuality between audio and visual information. To further obtain a better alignment of the latent space, we derive a new evidence lower bound (ELBO) with a Wasserstein distance~\cite{arjovsky2017wasserstein,deng2018latent,tian2019latent} constraint, which formulates our objective function.

The main contributions of our work can be summarized as follows:
1) We model the audio-visual correlations from the perspective of cross-modal generation. We make a shared-latent space assumption to apply a unified representation for two modalities, by deriving the objective function from a new lower bound with a Wasserstein distance constraint. 
2) We propose the \emph{MS-VAE} network, which is a novel self-supervised learning framework for the audio-visual cross-modal generation. \emph{MS-VAE} generates a corresponding audio-video pair from either single modality input and alleviates the degeneration problem in VAE training.
3) The learned latent representations from \emph{MS-VAE} can be readily applied in multiple audio-visual tasks. Experiments on AVE dataset~\cite{tian2018ave} show that our unsupervised method is able to achieve performance comparable or superior to the supervised methods on the challenging localization and retrieval tasks, even trained without any labels.

\section{Methodology}

\subsection{MS-VAE for Cross-Modal Learning}

Our \emph{MS-VAE} network is composed of separate encoders and one shared decoder. $q_{a,\phi_{a}}(z_{a}|x_{a})$ is the encoder that encodes audio input data $x_{a}$ into a latent space $z_{a}$, and $q_{v,\phi_{v}}(z_{v}|x_{v})$ is the encoder for visual input data $x_{v}$ to construct another latent space $z_{v}$. Ideally, we wish to obtain an aligned mutual latent space $z$ where $z = z_{a} = z_{v}$. The shared decoder $p_{\theta}(x|z)$ aims to reconstruct the original data pair $x$ from this mutual latent space $z$ in training, in which case the expected reconstructed data should consist of $x_{a}$ and $x_{v}$, we denote the pair of audio and video data as $x = (x_{a}, x_{v})$. $\phi_{a}$, $\phi_{v}$ and $\theta$ are the model parameters, which we omit in the following formulations to reduce redundancy.

The goal of our model resembles to the original VAE~\cite{kingma2013auto}, where we target to maximize the log-probability $\log p(x)$ of the reconstruction data pair $x$ from the desired mutual latent space $z$, $i$ represents either the modality $a$ (audio) or $v$ (visual). This model design leads to a similar variational lower bound as the original VAE~\cite{kingma2013auto} as follows:

\begin{equation}
    \centering
    \begin{split}
     \log p(x) & \geq 
     \mathbb{E}_{z_i \sim q_i(z_i|x_i)}[\log p(x|z_i)] - \operatorname{KL}(q_{i}(z_{i}|x_{i})||p(z_{i})),
     \label{eq:1}
    \end{split}
\end{equation}

\noindent where $\operatorname{KL}$ denotes the Kullback-Leibler divergence, defined as $\operatorname{KL}(p(x)||q(x)) = \int_{x}p(x)\log\frac{p(x)}{q(x)}$, which measures the similarity between two distributions and is always positive.
To build the relation between audio and video components, we rewrite Equation \ref{eq:1} as a mixture of log-likelihood conditioned on the latent variable from different modality. A Wasserstein distance loss~\cite{bonneel2015sliced}, which we refer as $\mathcal{W}_\mathrm{latent}$, is further added to better encourage the alignment between two latent space. Since the Wasserstein distance is always positive, the inequality remains valid. In this case, we obtain a new lower bound, whose equality is obtained only when the modeled distribution is the same as data distribution, as well as $z_{a}$ and $z_{v}$ are perfectly aligned:

\begin{equation}
\centering
\begin{split}
   & \log\,p(x)    \geq
    \mathop{{}\mathbb{E}_{z}}[\log(p(x|z)]  - \frac{1}{2} [\operatorname{KL}(q_{a}(z|x_{a})||p(z)) \\
    & + \operatorname{KL}(q_{v}(z|x_{v})||p(z))]
     - \mathcal{W}_\mathrm{latent}(q_{a}(z_a|x_{a})||q_{v}(z_v|x_{v})). \\
\end{split}
\label{eq:2}
\end{equation}

\subsection{Network and Training}

The schematic overview of the proposed \emph{MS-VAE} architecture is illustrated in Figure~\ref{fig:msvae}. We have separate encoders $q_a$ and $q_v$ for audio and visual inputs, a shared decoder $p$ is used to generate the corresponding audio and visual data pair. Wasserstein distance is computed between two latent variables $z_a$ and $z_v$ to encourage the alignment between the latent space using the similar approach as in WAE~\cite{tolstikhin2017wasserstein}, where we sample from latent variables $z_a$ and $z_v$ to compute $\mathbb{E}_{q_a,q_v}\left[||z_{a},z_{v}||_{2}\right]$.

The ultimate goal of the proposed \emph{MS-VAE} is to obtain an aligned latent representation that automatically learns the audio-visual correlations.
During training, for each epoch, we reconstruct the audio-visual pair from either audio or visual input. The encoder returns the $\mu_{i}$ and $\sigma_{i}$ for the Gaussian distribution, $z_i$ is sampled from $\mathcal{N}(\mu_{i},\sigma_{i})$. The reconstruction loss Mean Square Error (MSE) is computed between the reconstructed pair $\hat{x_i}$ from modality $i$ and the input ground truth pair $x$. The total loss contains the reconstruction loss, $\operatorname{KL}$ divergence and the Wasserstein latent loss. No label information is given in the entire training process. Overall speaking, we have three loss terms to optimize:

\begin{equation}
    \centering
    \mathcal{L}_\mathrm{total} = \lambda_{1}\mathcal{L}_{\mathrm{MSE}} + \lambda_{2}\mathcal{L}_{\operatorname{KL}} + \lambda_{3}\mathcal{W}_\mathrm{latent}.
\label{eq:loss}
\end{equation}

Empirically, we choose $\lambda_{1}$ and $\lambda_{3}$ to be 1, $\lambda_{2}$ is set to be 0.1 in the first 10 training epochs and then reduced to 0.01 to encourage better reconstruction.

\section{Experiments}

\subsection{Dataset and Evaluation Metrics}

We expect the data for our experiments have intrinsic correlations in describing an audio-visual event, we therefore adopt the AVE dataset. The AVE dataset~\cite{tian2018ave} is a subset of AudioSet~\cite{gemmeke2017audio} that contains 4,143 videos labeled with 28 event categories. Each video has a duration of 10s, and at least 2s are labeled with audio-visual events, such as baby crying and dog barking \textit{etc.}. Apart from the event categories labels for the entire video, each video is also temporally divided into 10 segments with audio-visual event boundaries, indicating whether the one-second segment is event-relevant or the background. The labels in \emph{MS-VAE} are purely used for evaluation purposes in our experiments, no labels are used in training.

We mainly apply our proposed model in two downstream tasks: the cross-modal localization (CML)~\cite{tian2018ave} and the audio-visual retrieval~\cite{arandjelovic2018objects}.
The CML task contains two subtasks, including localizing the visual event boundary from audio signals (A2V) and vice versa (V2A). The evaluation accuracy is computed based on the strict exact match, which means that we only count correct when the match location is exactly the same as its ground truth.
For the retrieval task, the mean reciprocal rank (MRR) is used to evaluate the performance. MRR calculates the average value of the reciprocal of the rank at which the first relevant information is retrieved across queries.
In addition to the concrete audio-visual tasks, we also include further qualitative ablation studies on the learned latent space to provide a more comprehensive analysis for the proposed \emph{MS-VAE} model.

\subsection{Cross-Modal Localization Task}

Cross-modal localization is proposed in~\cite{tian2018ave}, in which we wish to temporally locate a given segment of one modality (audio/visual) data in the entire sequence of the other modality. The localization is realized in two directions, \emph{i.e.}, visual localization from given audio segments (A2V) and audio localization from given visual segments (V2A). This task especially emphasizes the correlations between visual and audio signals since not only the correlations between different modalities are required, the temporal correlations are also needed to successfully fulfill the task~\cite{tian2018ave,wu2019DAM,lin2019dual,ramaswamy2020see}. 

In inference, for the sub-task A2V, we adopt the sliding window strategy as in~\cite{tian2018ave} to optimize the following objective: $t^* = \operatorname{argmin}_{t} \sum^{l}_{s=1} \mathcal{D}_{cml}(V_{s+t-1}, \hat{A}_{s})$, where $t^* \in \{ 1,...,T-l+1\}$ is the start time when audio and visual content synchronize, $T$ is the total length of a testing video sequence, and $l$ is the length of the audio query $\hat{A}_{s}$. The time position that minimize the cumulative distance between audio segments and visual segments is chosen to be the matching start position.
For $\mathcal{D}_{cml}$ in our experiments, we compute two terms, \emph{i.e.}, the Wasserstein latent distance $\mathcal{W}_{latent}$ between two latent variables encoded from audio and visual segments, and the Euclidean distance $\mathcal{D}_{gen}$ between the generated pair by audio and visual segment: $ \mathcal{D}_{cml} =  \mathcal{W}_{latent} +  \mathcal{D}_{gen}$. Similarly for another sub-task V2A.

\begin{table}[t]
\begin{center}
\scalebox{0.87}{
\begin{tabular}{|l|l|c|c|c|}
\hline
 Setting & Method & A2V $\uparrow$ & V2A $\uparrow$ & Average $\uparrow$ \\
\hline 
\multirow{4}{*}{Spv.} & DCCA~\cite{andrew2013deep} & 34.1 & 34.8 & 34.5 \\ 
& AVDLN~\cite{tian2018ave} & 35.6 & 44.8 & 40.2 \\ 
& AVSDN~\cite{lin2019dual} & 37.1 & 45.1 & 40.9 \\ 
& AVFB~\cite{ramaswamy2020see}& 37.3 & 46.0 & 41.6 \\ 
\hline
\multirow{2}{*}{Unspv.} & Ours & 25.0 $\pm$ 0.9 & 38.8 $\pm$ 0.4  & 31.9 $\pm$ 0.7\\
& Ours+$\mathcal{W}$ & 37.4 $\pm$ 1.2 & 40.0 $\pm$ 1.4 & 38.7 $\pm$ 1.3\\
\hline
\end{tabular}}
\end{center}
\caption{Quantitative evaluations on the CML task in terms of the exact matching accuracy. $Spv.$ means supervised and $Unspv.$ means unsupervised.}
\label{tab:cml}
\end{table}

\begin{figure}[t]
\begin{center}
\includegraphics[width=0.98\columnwidth]{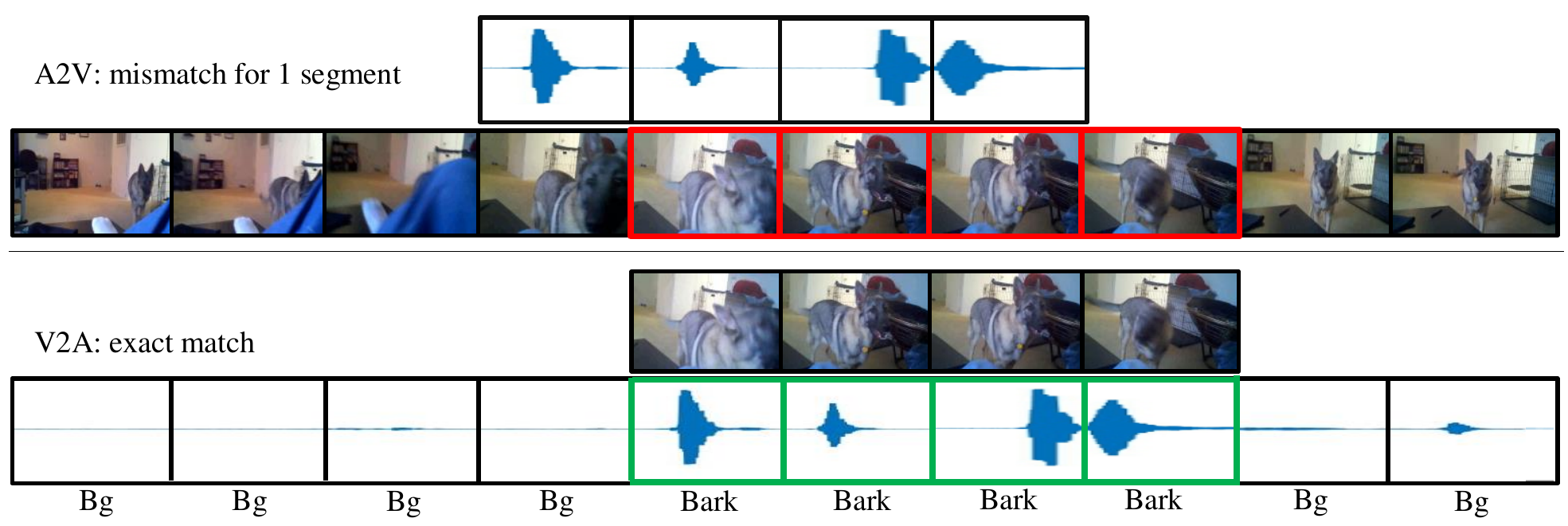}
\end{center}
\caption{Example of qualitative results for the CML task.}
\label{fig:cml}
\vspace{-0.2cm}
\end{figure}

The quantitative results are shown in Table~\ref{tab:cml}. We compare our method with other network models that learn the audio-visual correlation in a supervised manner. It is worth noting that our proposed self-supervised model achieves comparable performance with the supervised state-of-the-art methods.
Figure~\ref{fig:cml} shows an example of a mismatch and exact match for the cross-modal localization task. In A2V sub-task, we are given an audio query accompanying an event, \textit{e.g.,} dog bark, we want to localize its corresponding positions in the entire 10s visual frame sequence. Similarly in the V2A sub-task. Only the exact match is considered to be correct, thus making the task very challenging.

\subsection{Cross-Modal Retrieval Task}

\begin{table}[t]
\begin{center}
\scalebox{0.95}{
\begin{tabular}{|l|l|c|c|c|c|}
\hline
Setting& Method & A-A $\uparrow$ & A-V$\uparrow$ &V-A$\uparrow$ & V-V$\uparrow$ \\
\hline 
\multirow{3}{*}{Spv.}&AVDLN~\cite{tian2018ave} & 0.34 & 0.17 & 0.20 & 0.26 \\
&AVSDN~\cite{lin2019dual} & 0.38 & 0.19 & 0.21 &  0.25\\
&AVFB~\cite{ramaswamy2020see} & 0.37 & 0.20 & 0.20 & 0.27 \\ \hline
\multirow{5}{*}{Unspv.}&$L^3$~\cite{arandjelovic2017look} & 0.13 & 0.14 & 0.13 & 0.14 \\
&AVE~\cite{arandjelovic2018objects} & 0.16 & 0.14 & 0.14 & 0.16 \\ \cline{2-6}
&Ours & 0.24  & 0.14  &  0.15 & 0.27 \\ 
&Ours+$\mathcal{W}$ & 0.37 & 0.17 & 0.18 & 0.24  \\
\hline
\end{tabular}}
\end{center}
\vspace{-0.3cm}
\caption{Cross-modal and intra-modal retrieval results in terms of MRR. The columns headers denote the modalities of the query and the database. For example, A-V means retrieve visual frames from audio query. \textit{Spv.} and \textit{Unspv.} denote supervised and unsupervised.}
\label{tab:retrieval}
\vspace{-0.3cm}
\end{table}

Cross-modal retrieval task proposed in~\cite{arandjelovic2018objects} seeks to find the relevant audio/visual segment given a single audio or visual query. As in~\cite{arandjelovic2018objects}, we randomly sample a single event-relevant pair of visual frame and audio signal from each video in the testing set of AVE dataset to form the retrieval database. The retrieval task contains intra-modal (\emph{e.g.}, audio-to-audio) and cross-modal (\emph{e.g.}, audio-to-visual) categories. Each item in the test set is used as a query. The identical item from the database for each given query in intra-modal retrieval is removed to avoid bias. All the labels are only used for evaluations to calculate the MRR metric, which computes the average of the reciprocal of the rank for the first correctly retrieved items (\textit{i.e.}, items with the same category label as the query).

We fine-tune the $L^{3}$-Net~\cite{arandjelovic2017look} and AVE-Net~\cite{arandjelovic2018objects} in the training set of AVE dataset. AVE-Net incorporates a Euclidean distance between two separate feature embeddings to enforce the alignment. Both models learn audio-visual embedding in a self-supervised manner. We use the same inference as presented in~\cite{arandjelovic2017look} and~\cite{arandjelovic2018objects} for $L^{3}$-Net and AVE-Net, respectively. In addition to these two unsupervised baselines, we also adopt the supervised models to perform the retrieval tasks. Similar to the previous CML task, these models are trained to minimize the similarity distance between two embeddings of a corresponding audio-visual pair. In inference, we use a similar technique as in cross-modal localization task, which is to retrieve the sample that has the minimum distance score. For \emph{MS-VAE}, we compare the sum of Wasserstein latent distance and the reconstruction error between the given query and all the test samples from the retrieval database.

Table~\ref{tab:retrieval} presents the quantitative comparison between the proposed \emph{MS-VAE} and other methods. We achieve better performance in all four sub-tasks compared to the unsupervised baselines, and achieve competitive performance close to the supervised models.
This experiment further proves the effectiveness of leveraging audio-visual correlations for \emph{MS-VAE}. 

\subsection{Ablation Analysis}

\begin{figure}[t]
\begin{center}
\includegraphics[width=0.98\columnwidth]{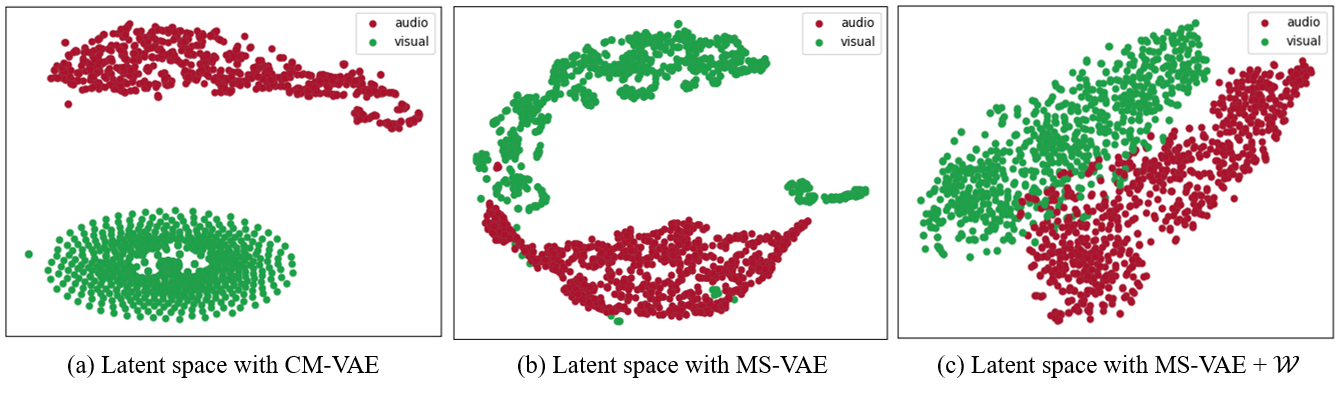}
\end{center}
\vspace{-0.5cm}
\caption{Latent space comparison. t-SNE visualization for testing segments of corresponding audio-visual pair.}
\label{fig:latent}
\vspace{-0.3cm}
\end{figure}

We perform additional ablation studies on the shared decoder and the Wasserstein restriction components to show that they are both significant factors that contribute to the learned audio-visual correlations.
Figure~\ref{fig:latent} shows the qualitative results for the latent space comparison among \emph{CM-VAE}~\cite{spurr2018cross}, our \emph{MS-VAE} and \emph{MS-VAE} + $\mathcal{W}$ in the form of latent space visualization. Specifically, \emph{CM-VAE} is proposed in~\cite{spurr2018cross} for cross-modal generation for hand poses (\textit{e.g.}, generating 3D hand poses from RGB images). \emph{CM-VAE} adopts separate encoders and decoders for each modality, and can be considered as \emph{MS-VAE} without shared decoder version.
It is interesting to observe that in the latent space obtained by \emph{CM-VAE}, the visual embedding degenerates into a purely non-informative Gaussian distribution as in~\cite{zheng2018degeneration}. \emph{MS-VAE} alleviates the degeneration problem and learns similar distributions for audio and visual input, but the distance between two latent space is still evident. Wasserstein distance further bridges the two latent space and regularizes the learned data distributions. Note that the perfect alignment is very difficult to achieve and remains to be an open question in the research community.

\section{Conclusion}
In this paper, we propose the \emph{MS-VAE} framework to learn the intrinsic audio-visual correlations for multiple downstream audio-visual tasks. \emph{MS-VAE} is a self-supervised learning framework that generates a pair of corresponding audio-visual data given either one modality as input data. 
It leverages the advantage of label-free self-supervised learning from the generative models and achieves very competitive performance for multiple audio-visual tasks. 

\section{Acknowledgements}
This research was partially supported by NSF NeTS-1909185, CSR-1908658 and Cisco. This article solely reflects the opinions and conclusions of its authors and not the funding agents. Yu Wu is supported by the Google PhD Fellowship.

\bibliographystyle{IEEEbib}
\bibliography{strings,refs}

\end{document}